\documentclass[a4paper]{article}

\usepackage{INTERSPEECH2019}
\usepackage{amssymb,amsmath,graphicx,epsfig,color,cite}
\usepackage[english,capitalize]{cleveref}

\title{Auxiliary Interference Speaker Loss for Target-Speaker Speech Recognition}
\name{Naoyuki Kanda$^1$, Shota Horiguchi$^1$, Ryoichi Takashima$^1$, Yusuke Fujita$^1$, Kenji Nagamatsu$^1$, \\Shinji Watanabe$^2$}
\address{
  $^1$Hitachi Ltd., Japan\\
  $^2$Johns Hopkins University, USA}
\email{\{naoyuki.kanda.kn, shota.horiguchi.wk, ryoichi.takashima.dh, yusuke.fujita.su, kenji.nagamatsu.dm\}@hitachi.com, shinjiw@ieee.org}

\begin{document}

\maketitle
\begin{abstract}
In this paper, we propose a novel auxiliary loss function for target-speaker automatic speech recognition 
(ASR). Our method
automatically extracts and transcribes target speaker's utterances from a monaural mixture of multiple speakers speech 
given a short sample of the target speaker. 
The
proposed auxiliary loss function attempts to additionally maximize interference speaker ASR accuracy during training. 
This will regularize the network to achieve a better representation for speaker separation, 
thus achieving better accuracy on the target-speaker ASR. 
We evaluated our proposed method using two-speaker-mixed speech in various signal-to-interference-ratio conditions. 
We first built a strong target-speaker ASR baseline 
based on the state-of-the-art lattice-free maximum mutual information.
This baseline achieved
a word error rate (WER) of 18.06\% on the test set 
while a normal ASR trained with clean data produced a completely corrupted result (WER of 84.71\%). 
Then, our proposed loss 
further reduced the WER by 6.6\% relative to 
this strong 
baseline, 
achieving a WER of 16.87\%. 
In addition to the accuracy improvement, 
we also showed that 
the auxiliary output branch for the proposed loss can even be used
for a secondary ASR for interference speakers' speech. 
\end{abstract}

\section{Introduction}
Thanks to the recent advances in deep-learning \cite{seide2011conversational,dahl2012context,hinton2012deep}, 
the accuracy of automatic speech recognition (ASR) for some datasets have become 
close to (e.g., Switchboard \cite{xiong2016achieving,saon2017english,xiong2017toward}) or 
beyond (e.g., LibriSpeech in \cite{amodei2016deep} and \cite{kanda-interspeech2018}) the level of human transcribers. 
However, despite this progress, the accuracy of multi-talker speech recognition 
is still very limited \cite{yoshioka2018recognizing,barker2018fifth,kanda2018hitachi} 
because of the difficulty of separating the target speaker's speech from other speech. 
One example is meeting speech recognition, which is known for having word error rates (WERs) 
around 30\% (e.g. \cite{yoshioka2018recognizing,kanda2019icassp}) even with state-of-the-art speech recognizers. 

In this paper, we focus on ASR for monaural audio that contains overlapped speech uttered by multiple speakers. 
One direction to solve this problem is applying a speech separation method before ASR, 
such as deep clustering \cite{hershey2016deep}, deep attractor network \cite{chen2017deep}, etc. 
However, a major drawback of this approach is that the training criteria for speech separation do not necessarily maximize ASR accuracy. 
If the goal is ASR, it will be better to use training criteria that directly maximize ASR accuracy.

Recently, multi-speaker ASR based on permutation invariant training (PIT) 
was proposed \cite{yu2017recognizing,chen2018progressive,settle2018end,seki2018purely,chang2019end}. 
In the PIT scheme, the label permutation problem is solved by considering all possible permutations 
when calculating the loss \cite{yu2017permutation}. 
PIT was first proposed for speech separation \cite{yu2017permutation} and soon extended to 
ASR loss with promising results \cite{yu2017recognizing,chen2018progressive,settle2018end,seki2018purely,chang2019end}. 
However, one hidden drawback of PIT arises when speaker tracing or speaker diarization is required. 
Namely, a PIT-ASR model produces transcriptions {\it for each utterance} of speakers in an unordered way, 
and it is no longer straightforward to solve speaker permutations {\it across utterances}.
To make things worse, 
a PIT model trained with ASR-based loss normally does not produce separated speech waveforms, 
which makes speaker tracing more difficult. 
For applications in which speaker tracing or speaker diarization has an essential role (e.g. ASR for meeting recordings), 
it could become a serious drawback of the PIT model. 

On the other hand, target-speaker ASR, which automatically extracts and transcribes only the target speaker's utterances 
given a short sample of the target speaker's speech, has been proposed \cite{zmolikova2017speaker,delcroix2018single}. 
Zmolikova et al. proposed a target-speaker neural beamformer that extracts a target speaker's utterances 
given a short sample of the target speaker's speech \cite{zmolikova2017speaker}. 
This model was recently extended to deal with ASR-based loss to maximize ASR accuracy with promising results \cite{delcroix2018single}. 
While the target-speaker models require additional input of a target speaker's speech sample, 
it can naturally solve the speaker permutation problem {\it across utterances} 
without using additional speaker identification after ASR. 
We believe this property of target-speaker ASR is attractive for many practical applications. 

\begin{figure}[t]
\centerline{\epsfig{figure=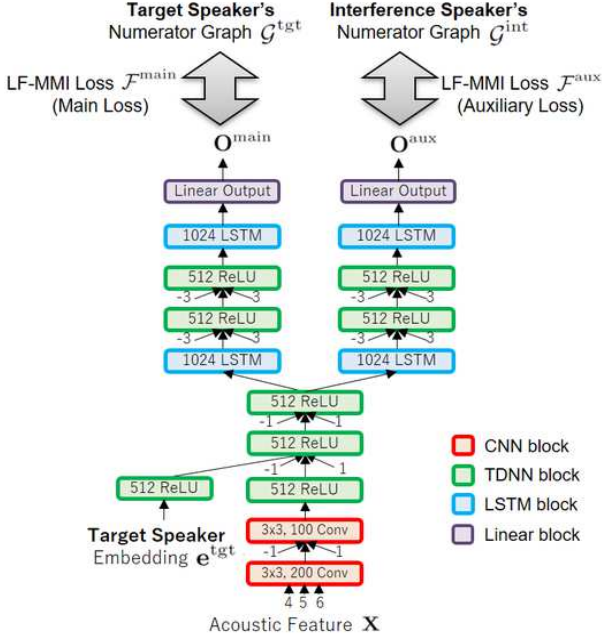,width=80mm}}
%\centerline{\includegraphics[width=80mm]{overview7.pdf}}
\vspace{-3mm}
\caption{{\it Overview of target-speaker AM architecture with auxiliary interference speaker loss. Auxiliary networks for interference speaker loss are only used in training and normally removed in the decoding phase. A number with an arrow indicates a time splicing index, which forms the basis of a time-delay neural network (TDNN) \cite{peddinti2015time}. The input features were advanced by five frames, which has the same effect as reference label delay.}}
\label{fig:overview}
\vspace{-3mm}
\end{figure}

Following the discussion above, in this paper, we focus on target-speaker ASR, 
and propose a novel auxiliary loss function to improve target-speaker ASR accuracy. 
Our proposed loss function attempts to additionally maximize interference speaker ASR accuracy (\cref{fig:overview}) 
and will regularize the network to achieve better representation, which, as a result, produces better target-speaker ASR accuracy 
in multi-speaker conditions. 
We evaluated our proposed method using two-speaker-mixed speech in various signal-to-interference (SIR)-ratio 
conditions to demonstrate its effectiveness. 
We also conducted various investigations on our proposed method and model architectures,
including the possibility to use
the auxiliary branch for the proposed loss 
for a secondary ASR. 
In this secondary ASR setting, 
our model can explicitly output 
not only transcriptions of the target speaker but also those of other speakers
in a consistent order {\it across utterances},
which is another unique property of our model.

As an additional contribution of this paper, to the best of our knowledge, 
this is the first report applying a lattice-free maximum mutual information (LF-MMI)-based acoustic model (AM) \cite{povey2016purely} 
for target-speaker ASR\footnote{LF-MMI has been applied for the PIT-ASR model \cite{chen2018progressive}.}.
Thanks to the state-of-the-art performance given by LF-MMI, our results were fairly good. 
For example, we achieved a WER of 16.50\% for the ``wsj0-2mix'' dataset \cite{hershey2016deep} for which 
WERs in the range of 25 - 30\% 
have been reported \cite{isik2016single,qian2018single,seki2018purely,chang2019end}. 

\section{Auxiliary Interference Speaker Loss} 
\label{sec:propose}
In this section, we explain our proposed method and its use with an LF-MMI-based AM 
due to its state-of-the-art performance \cite{povey2016purely,kanda2017investigation,kanda2018sequence}. 
However, it should be noted that our work can be extended to any kind of ASR loss 
like cross entropy (CE) \cite{seide2011conversational}, state-level minimum Bayes risk (sMBR) \cite{vesely2013sequence,su2013error}, 
and lattice-free sMBR \cite{kanda-interspeech2018}. 
In addition, the idea can even be extended to end-to-end models 
like connectionist temporal classification (CTC) \cite{graves2006connectionist} and 
attention encoder-decoder-based models \cite{chorowski2014end,chan2016listen}.

\subsection{Overview of the proposed model}
\cref{fig:overview} describes our proposed model architecture for the LF-MMI AM. 
The network has two input branches. One branch accepts acoustic features ${\bf X}$ as a normal AM 
while the other branch accepts an embedding ${\bf e}^{\rm tgt}$ that represents the characteristics of the target speaker. 
In this study, we used a log Mel-filterbank (FBANK) and i-vector \cite{dehak2011front,saon2013speaker} for the acoustic features 
and target speaker embedding, respectively. 

The key idea is in its output branch. 
The proposed model has multiple output branches which produce outputs ${\bf O}^{\rm main}$ and ${\bf O}^{\rm aux}$ for the main and proposed auxiliary loss functions, respectively. 
The main loss attempts to maximize the target speaker's ASR accuracy, 
while the auxiliary loss attempts to maximize interference speaker ASR accuracy. 
In this study, we used LF-MMI for both criteria, details of which are explained at the latter section. 
Our assumption is as follows: By maximizing the ASR accuracy of the interference speaker as well as that of the target speaker, 
a better representation for speaker separation is learned in the shared layers of the network, 
resulting in improved accuracy on the target speaker's ASR. 

The network is trained with a mixture of multi-speaker speech given their transcriptions. 
We assume that, for each training data, (1) at least two speakers' transcriptions are given, 
(2) at least the transcription for the target speaker is marked so that we can identify the target speaker's transcription, 
and (3) a sample for the target speaker can be used to extract speaker embeddings. 
This assumption can be easily satisfied by artificially generating training data by mixing multiple speakers' speech. 
After finishing the network training, the auxiliary output branches corresponding to the interference speakers are removed, 
and only the network branch for the target speaker is used for the decoding.

\subsection{Main loss function}
In the case of LF-MMI, the main loss function (to minimize) for the target speaker is 
defined as \footnote{This is a numerator-graph ($\mathcal{G}_u^{\rm tgt}$)-based extension of 
a basic MMI-criterion $\mathcal{F}^{\mathit{MMI}}=\sum_{u} -\log P({\bf S}_u|{\bf O}^{\rm main}_{u})$ \cite{vesely2013sequence,su2013error}, 
which uses a Viterbi-aligned reference state sequence ${\bf S}_u$ instead of $\mathcal{G}^{\rm tgt}_u$.}
\begin{align}
\mathcal{F}^{\rm main}&=\sum_{u} {\mathrm{LFMMI}}({\bf O}_{u}^{\rm main};\mathcal{G}^{\rm tgt}_{u}),  \nonumber \\
=&\sum_{u} \sum_{{\bf S}} -P({\bf S}|{\bf O}_{u}^{\rm main},\mathcal{G}^{\rm tgt}_{u}) \log P({\bf S}|{\bf O}_{u}^{\rm main},\mathcal{G}^{D}), \nonumber \label{eq:mmi}
\end{align}
where $u$ is the index of training samples. %utterances. 
The term $\mathcal{G}^{\rm tgt}_u$ indicates a numerator (or reference) graph that 
represents a set of possible correct state sequences for the utterance of the target speaker of $u$-th training sample. 
The term ${\bf S}$ denotes a hypothesis state sequence for  $u$-th training sample. %${\bf X}_u$. 
The term $\mathcal{G}^{D}$ denotes a denominator graph, which represents a possible hypothesis space and 
normally consists of a 4-gram phone language model in LF-MMI training \cite{povey2016purely}.

Note that one important technique known as cross entropy (CE)-regularization is normally used 
for LF-MMI training \cite{povey2016purely}. 
With this technique, an additional output layer that is optimized on the basis of the CE criterion is introduced. 
In training, parameters are optimized to minimize the weighted interpolation of LF-MMI and CE criteria. 
We used CE-regularization for our evaluation not only for the main loss but also for the auxiliary interference speaker loss. 
However, we omit the CE-regularization term here for the brevity of the equations.

\begin{table*}[t]
\caption{\label{tab:compare-loss} {\it WERs (\%) for the two-speaker mixed LibriSpeech corpus in various SIR conditions. Note that for clean single-speaker speech, a clean AM achieved WERs of 4.88\% and 5.54\% for dev-clean and test-clean, respectively.}}
\vspace{-3mm}
\centering
{\footnotesize
\begin{tabular}{c|cccccc|cccccc} \hline
        & \multicolumn{6}{c|}{dev-clean (two spkeakers mixed)} & \multicolumn{6}{c}{test-clean (two spkeakers mixed)}\\ 
SIR of the targeted speaker's speech    & 10 & 5  & 0 & -5 & -10 & Avg. & 10 & 5  & 0 & -5 & -10 & Avg.\\ \hline
Clean AM                             & 65.14 & 78.72 & 88.56 & 91.96 & 94.31 & 83.74 & 66.92 & 79.45 & 89.18 &92.87  & 95.15 & 84.71 \\
Target-Speaker AM w/o aux. loss    & 13.98 & 15.02 & 16.80 & 18.42 & 21.54 & 17.15 & 15.59 & 16.13 & 17.65 & 19.17 & 21.78 & 18.06 \\
Target-Speaker AM w/ aux. loss   & {\bf 13.51} & {\bf 14.32} & {\bf 16.02} & {\bf 17.57} & {\bf 20.51} & {\bf 16.39} & {\bf 14.59}& {\bf 15.06} &{\bf 16.46} & {\bf 17.86} & {\bf 20.38} & {\bf 16.87} \\ \hline
\end{tabular}
}
\vspace{-3mm}
\end{table*}

\subsection{Auxiliary interference speaker loss}
The proposed auxiliary interference speaker loss is defined to maximize interference speaker ASR accuracy, 
which we expect to enhance the speaker separation ability of the network. 
In the case of using the LF-MMI criterion, the loss is defined as follows\footnote{The loss can be extended 
in the case of multiple interference speakers in accordance with the PIT principle, as follows.
\begin{align}
\mathcal{F}^{\rm aux}=\sum_{u} \min_{i\in {\rm permu}(I)} \sum_{n} {\mathrm{LFMMI}}({\bf O}_{u,n}^{\rm aux};\mathcal{G}^{\rm int}_{u,i[n]}),  \nonumber 
\end{align}
where ${\rm permu}(I)$ 
represents 
a set of permutations of interference speakers $\{1,...,I\}$, and
$n$ denotes the index of the permutation $i$.} 
\begin{align}
\mathcal{F}^{\rm aux}=\sum_{u} {\mathrm{LFMMI}}({\bf O}_{u}^{\rm aux};\mathcal{G}^{\rm int}_{u}),  \nonumber 
\end{align}
where $\mathcal{G}^{\rm int}_u$ denotes a numerator (or reference) graph that represents a set of possible 
correct state sequences for the utterance of the interference speaker of $u$-th training sample.

Finally, the loss function $\mathcal{F}^{\rm comb}$ for training is defined as the combination of the main and auxiliary losses, as
\begin{align}
\mathcal{F}^{\rm comb}=\mathcal{F}^{\rm main}+\alpha\cdot\mathcal{F}^{\rm aux}, \nonumber
\end{align}
where $\alpha$ is the scaling factor for the auxiliary loss. 
In our evaluation, we set $\alpha=1.0$, which gave us promising results.

Note that the original target-speaker ASR corresponds to the model without the proposed auxiliary loss; 
Namely, it is the case of $\alpha=0.0$.

\section{Evaluation}
\label{sec:eval}
\subsection{Experiment with LibriSpeech}
\subsubsection{Experimental settings}
For our primary evaluation, we used the LibriSpeech corpus \cite{panayotov2015librispeech}, 
which consists of about 1,000 hours of read-aloud English speech. 
In this study, we used 100 hours of the clean portion of the dataset for training the AMs. 
For evaluation, we used ``dev-clean'' and ``test-clean'' in accordance with 
the Kaldi recipes \cite{povey2011kaldi} as the basic materials for development and evaluation sets, respectively.

For all training, development, and evaluation data, we artificially generated two-speaker mixed speech in accordance with the protocol below.
\begin{enumerate}
\item Prepare a list of speech samples (main list), which is the main target of ASR.
\item Shuffle the main list to create a second list under the constraint 
that the same speaker does not appear in the same line in the main and second lists.
\item Mix the audio in the main list and the second list one-by-one, with a specific SIR. 
For training data, we randomly sampled an SIR value from uniform distribution between -10 dB and 10 dB for each mixture. 
For the development and evaluation data, we generated data with an SIR of \{10, 5, 0, -5, -10\} dB.
\item Only in the case of the training data, the volume of each mixed speech was randomly changed to enhance robustness for the volume difference.
\end{enumerate}
Note that, in accordance with the protocol above, the speech of the target speaker could be much shorter or 
much longer than that of the interference speaker. We intentionally selected this protocol because we believe it is also important to evaluate the ability to correctly select the portion of the target speaker's speech when a single speaker, 
who could be either target speaker or interference speaker, is speaking. 

We trained an acoustic model consisting of a convolutional neural network (CNN), 
time-delay neural network (TDNN) \cite{waibel1989phoneme}, and long short-term memory (LSTM) \cite{hochreiter1997long} 
as shown in \cref{fig:overview}. 
Input acoustic features for the network were 40-dimensional log-Mel-filterbank (FBANK) without normalization. 
In addition, a 100-dimensional i-vector was extracted and used for the target speaker embedding to indicate the target speaker. 
For extracting i-vector, we randomly selected an utterance of the same speaker. 
We conducted 8 epochs of training on the basis of LF-MMI, where the initial learning rate was set to 0.001 
and exponentially decayed to 0.0001 by the end of the training. 
We applied $l2$-regularization and CE-regularization \cite{povey2016purely} with scales of 0.00005 and 0.1, respectively. 
The leaky hidden Markov model coefficient was set to 0.1. 
In addition, a backstitch technique \cite{wang2017backstitch} with a backstitch scale of 1.0 and backstitch interval of 4 was used.

For comparison purposes, we trained the AM without the proposed auxiliary loss, 
which corresponds to the original target-speaker model. 
We also trained a ``clean AM'' using clean, non-speaker-mixed speech. 
For this model, we also used a model architecture without the auxiliary output branch, and i-vector was extracted every 100 msec 
to realize online speaker/environment adaptation\footnote{This is one of the standard use cases for a clean AM in the Kaldi toolkit.}.

In decoding, we used an officially provided large 4-gram LM. 
For each test utterance, a sample of the target speaker was randomly selected 
from the other utterances in the test set\footnote{We used the same random seed over all experiments so that 
all experiments were conducted on the basis of the same conditions.}.
The average duration of the target speaker's sample was 7.2 sec and 7.4 sec for dev-clean and test-clean, respectively. 
All decoding parameters were tuned by the development set, and the best parameters were used for the evaluation set. 
All of our experiments were conducted on the basis of the Kaldi toolkit \cite{povey2011kaldi}. 

\subsubsection{Effect of auxiliary interference speaker loss}
The first row of \cref{tab:compare-loss} shows the result for the AM trained with clean data (``clean AM''). 
Note that WERs for clean speech with the clean AM was 4.88\% and 5.54\% for dev-clean and test-clean, respectively. 
As shown in the table, WERs were severely degraded by mixing the two speakers' speech, and the clean AM produced 
WER averages of 83.74\% and 84.71\% for dev-clean and test-clean, respectively.

The second row of \cref{tab:compare-loss} shows the result for the original target-speaker AM without auxiliary loss. 
This model dramatically improved the accuracy, achieving WER averages of 17.15\% and 18.06\% for dev-clean and test-clean, respectively. 

The third row of \cref{tab:compare-loss} shows the result for the proposed auxiliary loss function. 
It achieved the best WER among all SIR conditions. 
The proposed auxiliary loss achieved WERs of 16.39\% and 16.87\% for dev-clean and test-clean, respectively; a relative WER reduction of 6.6\%. 

\subsubsection{Comparison of model architectures}
\begin{figure}[t]
\centerline{\epsfig{figure=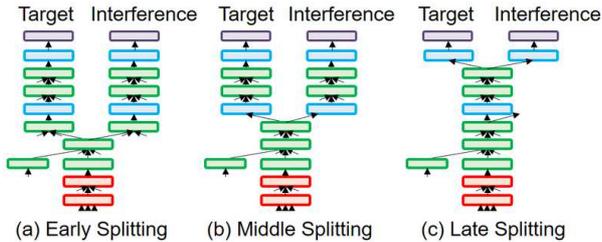,width=80mm}}
%\centerline{\includegraphics[width=80mm]{models.pdf}}
\vspace{-3mm}
\caption{{\it Model architectures with early, middle, and late splitting. Note that the middle splitting model was used as the default model in other experiments.}}
\label{fig:models}
\vspace{-3mm}
\end{figure}

To better understand the proposed method, we investigated the effect of the model architecture for the auxiliary loss. 
Although we added the auxiliary output branch at the middle point of the network in \cref{fig:overview}, 
we could add the auxiliary output branch at a different point as shown in \cref{fig:models}.

\begin{figure}[t]
\centerline{\epsfig{figure=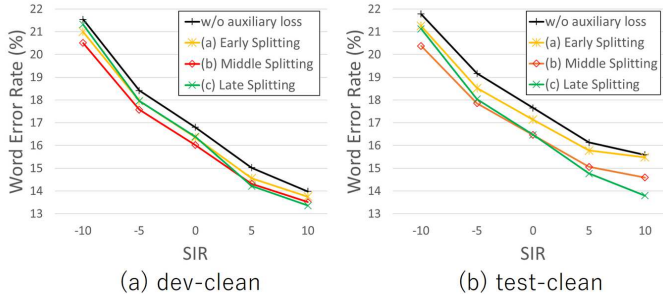,width=90mm}}
%\centerline{\includegraphics[width=90mm]{model-compare-result5.pdf}}
\vspace{-3mm}
\caption{{\it Comparison among model architectures in \cref{fig:models}.}}
\label{fig:model-compare-result}
\vspace{-3mm}
\end{figure}

The results for each model architecture are shown in \cref{fig:model-compare-result}. 
We first found that the early splitting (\cref{fig:models} (a)) was ineffective 
in all cases. 
We believe this to be natural because if we split the network early, 
only a limited number of network parameters can be used for speech separation.

The comparison of the middle splitting (\cref{fig:models} (b)) and the late splitting (\cref{fig:models} (c)) 
 produced some complicated results. 
As shown in \cref{fig:model-compare-result}, the middle splitting model showed 
robust improvements over broad SIR conditions. 
On the other hand, the late splitting model showed very good WERs when the SIR was higher than 5 dB, 
but the improvement became marginal when the SIR became low. 
Our interpretation is as follows. In the case of the late splitting model, 
the responsibility for the output prediction for the target speaker was concentrated on the last layer. 
Although this did not results in any severe problem when the evaluation condition was simple (i.e. a high SIR), 
problems did occur when the condition became more complicated. 
On the basis of these interpretation, we concluded that the middle splitting model is the most robust architecture.

\subsubsection{Interference speaker's ASR by the auxiliary network}

So far, we explained that the auxiliary output branch is removed in decoding. 
However, the auxiliary output branch could be used for interference speaker ASR. 
One possible scenario is that there is one target speaker whose 
utterance should be recognized with a mark of ``presenter'', and 
other audiences' utterances are recognized via the auxiliary branch with a mark of ``questioners.''

Therefore, we evaluated the ability of an auxiliary network for the secondary ASR.
In this evaluation, we provided the target speaker's embeddings for the network, 
and evaluated the WERs of the ASR results using the output of the auxiliary output branch. 
The results are shown in \cref{tab:second}. 
From this table, we found that the auxiliary output branch worked very well for the secondary ASR. 
This result clearly indicated that the shared layers of the network were learned to realize speaker separation as we expected.
In addition, we believe this secondary ASR itself is attractive as exemplified above.
Different from PIT-ASR models, 
our model can explicitly output 
not only transcriptions of the target speaker but also those of other speakers
in a consistent order {\it across utterances}.

\begin{table}[t]
\caption{\label{tab:second} {\it WERs (\%) for two-speaker-mixed test-clean. Main output branch was used for the target speaker's ASR and auxiliary output branch was used for the interference speaker's ASR.}}
\vspace{-3mm}
\centering
{\footnotesize
\begin{tabular}{cc|cc} \hline
SIR of  &  SIR of  & \multicolumn{2}{c}{WER (\%)} \\
 target spk.  &  interference spk. & target spk. & interference spk. \\  \hline
10 & -10     & 14.59 & 26.22 \\
5 & -5       & 15.06 & 19.90 \\
0 & 0      & 16.46 & 16.23 \\
-5 & 5     & 17.86 & 15.05 \\
-10 & 10    & 20.38 & 14.50 \\ \hline
\multicolumn{2}{c|}{Avg.} & 16.87 & 18.38 \\ \hline
\end{tabular}
}
\vspace{-3mm}
\end{table}

\subsection{Experiment with Wall Street Journal corpus}
For our final evaluation, we conducted experiments on the Wall Street Journal (WSJ) corpus. 
In accordance with \cite{chang2019end}, we used the WSJ SI284 for the training data and Dev93 for the development data. 
For evaluation, we used Eval92 and ``wsj0-2mix'' \cite{hershey2016deep}. 
When testing Dev93 and Eval92, we mixed two speakers' speech with SIRs of 0 dB. 
Other experimental settings were the same with those in the previous section.

The results are shown in \cref{tab:wsj}. We observed a significant improvement for Dev93 and moderate improvement for wsj0-2mix, 
while a marginal degradation of WER was observed for EVAL92. 
Note that the results by our AM were fairly good thanks to the state-of-the-art accuracy given by the LF-MMI. 
Our result for ``wsj0-2mix'' (16.50\% WER) is significantly better than the results reported by the conventional PIT-based ASR method, 
which achieved about a WER in the range of 25 - 30\% \cite{isik2016single,qian2018single,seki2018purely,chang2019end}.
 It is also significantly better than the result reported in the target-speaker ASR paper \cite{delcroix2018single}, 
which reported a WER of 34.0\% for the WSJ corpus in 0-dB-mixture settings. 

\begin{table}[t]
\caption{\label{tab:wsj} {\it WERs (\%) for WSJ corpus with clean AM and target-speaker (TS) AMs.}}
\vspace{-3mm}
\centering
{\footnotesize
\begin{tabular}{cccc} \hline
Model  &  Dev93  & Eval92  & wsj0-2mix \\ \hline
Clean AM  & 77.51  &78.03  & 79.81 \\
TS-AM w/o aux. loss & 12.24 & {\bf 11.32} & 16.78 \\
TS-AM w/ aux. loss & {\bf 11.31} & 11.38 & {\bf 16.50} \\ \hline
\end{tabular}
}
\vspace{-3mm}
\end{table}

\section{Conclusions}
\label{sec:conclusion}
In this paper, we proposed a novel auxiliary loss function for target-speaker ASR, 
in which it attempts to maximize interference speaker ASR accuracy. 
We evaluated our proposed method using two-speaker-mixed speech in various SIR conditions. 
We first built a strong target-speaker ASR baseline 
based on the state-of-the-art LF-MMI, achieving
a WER of 18.06\% on the test set 
while a normal ASR trained with clean data produced a completely corrupted result (WER of 84.71\%).
Then, our proposed loss 
further reduced the WER by 6.6\% relative to 
this strong target-speaker ASR baseline,
achieving a WER of 16.87\%. 
By investigating model architectures for the proposed loss, 
we determined that adding an auxiliary output branch from the middle point of the network worked the most robustly. 
We also showed that 
the auxiliary output branch can be used
for a secondary ASR for interference speakers' speech. 

\bibliographystyle{IEEEtran}
\bibliography{thesis}

\end{document}